\pdfoutput=1 
\documentclass[10pt,twocolumn,letterpaper]{article}

\usepackage{cvpr}              

\usepackage{graphicx}
\usepackage{amsmath}
\usepackage{amssymb}
\usepackage{booktabs}
\usepackage{multirow}
\usepackage{pifont}
\usepackage{xspace}

\usepackage[pagebackref,breaklinks,colorlinks]{hyperref}

\usepackage[capitalize]{cleveref}
\crefname{section}{Sec.}{Secs.}
\Crefname{section}{Section}{Sections}
\Crefname{table}{Table}{Tables}
\crefname{table}{Tab.}{Tabs.}

\def\cvprPaperID{8546} 
\def\confName{CVPR}
\def\confYear{2023}

\newcommand{\xmark}{\ding{55}}%
\newcommand{\cmark}{\ding{51}}%
\begin{document}

\title{Prompt Algebra for Task Composition }

\author{Pramuditha Perera\\
AWS AI Labs\\
{\tt\small pramudi@amazon.com}
\and
Matthew Trager\\
AWS AI Labs\\
{\tt\small mttrager@amazon.com}
\and
Luca Zancato\\
AWS AI Labs\\
{\tt\small zancato@amazon.it}
\and
Alessandro Achille\\
AWS AI Labs\\
{\tt\small aachille@amazon.com}
\and
Stefano Soatto\\
AWS AI Labs\\
{\tt\small soattos@amazon.com}
}
\maketitle

\begin{abstract}
We investigate whether prompts learned independently for different tasks can be later combined through  \textit{prompt algebra} to obtain a model that supports composition of tasks.  We consider Visual Language Models (VLM) with prompt tuning as our base classifier and formally define the notion of \textit{prompt algebra}.  We propose constrained prompt tuning to improve performance of the composite classifier. In the proposed scheme, prompts are constrained to appear in the lower dimensional subspace spanned by the basis vectors of the pre-trained vocabulary. Further regularization is added to ensure that the learned prompt is grounded correctly to the existing pre-trained vocabulary. We demonstrate the effectiveness of our method on object classification and object-attribute classification datasets. On average, our composite model obtains classification accuracy within 2.5\% of the best base model. On UTZappos it improves classification accuracy over the best base model by 8.45\% on average.
\end{abstract}

\section{Introduction}
\label{sec:intro}

\begin{figure}[t] 
\centering
\includegraphics[width=0.4\textwidth]{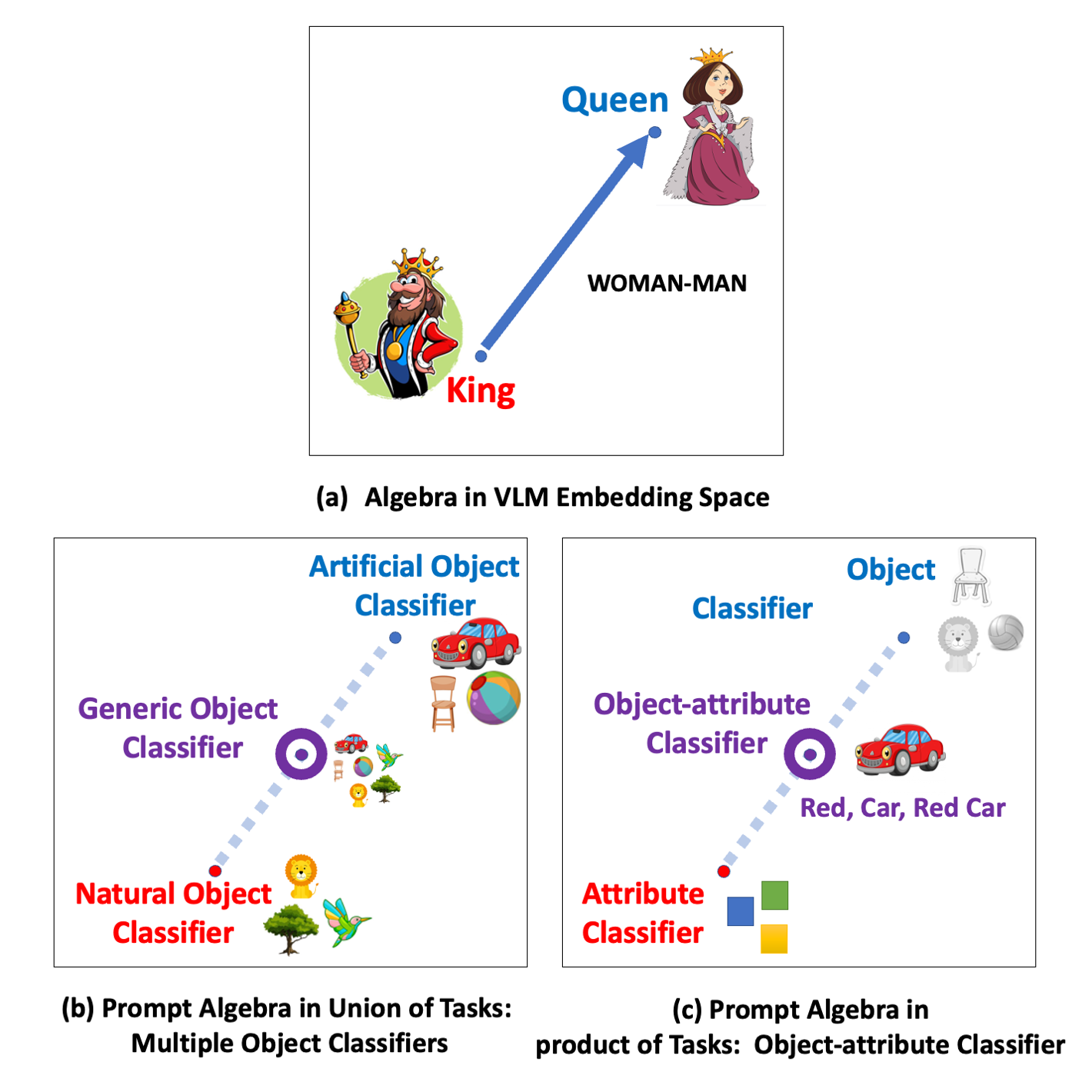} 
\caption{ Overview of prompt algebra. (a) VLM models exhibit compositionality between entities in the text feature space. (b) Prompt algebra can be used to combine prompts obtained from natural object classification and artificial object classification to produce prompts that result in a generic object classifier. (c) Combining attribute and object classifiers results in a prompt that performs well in object, attribute and object-attribute classification tasks.}
\label{fig:algebra}
\end{figure}

Recent works have shown that pre-trained text tokens respect compositional relationships in the text feature space between entities \cite{word2vec,tewel2021zero,distil}. For example, a text feature of \textit{Queen} can be obtained by adding the residual text feature of \textit{woman-man} to the text feature of \textit{King} as shown in Figure~\ref{fig:algebra} (a). On the other hand, the prompt-tuning literature \cite{lester-etal-2021-power, xianglisaliprefixtuning, liupretrainprompt, liu-etal-2022-p} suggests that usage of soft-prompt tokens to prompt a model can result in powerful classifiers that are on par with the classification ability of fine-tuned models. 

In this paper, we raise the question: are soft-prompt tokens compositional? Is it possible to combine two prompt-tokens that are trained independently on separate classification tasks to obtain a composite classifier (Figure~\ref{fig:algebra}b)? In our experiments, we provide evidence that such a relationship exists. We empirically show that prompt tokens can be manipulated using \textit{prompt algebra} to obtain classifiers for composite tasks.

This observation is important to the machine learning community for a simple reason. Recent advancements in deep learning\cite{he15deepresidual,WRN,vit} have paved the path for models with very competitive performance in specific downstream classification tasks. However, these models require annotated data for every class and have very limited flexibility to adapt to new tasks without retraining. Task composition enables composing two or more base classifiers to support classification in the full product space of classes avoiding the 
combinatorial explosion of classes. In general there are two forms of task compositionality that we are interested in:\\





\begin{figure}[t] 
\centering
\includegraphics[width=0.45\textwidth]{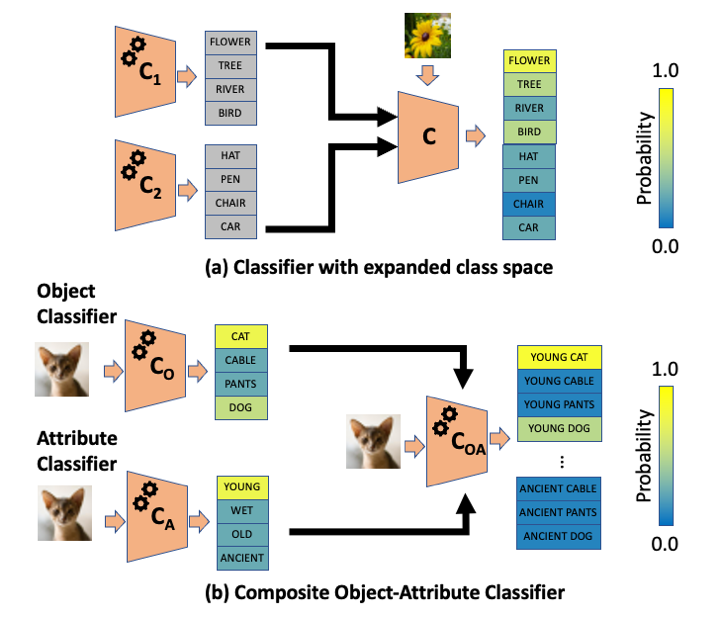} 
\caption{ Desired properties of a composed classifier. (a) The composed classifier should support the union of the classes supported by independently trained base classifiers. (b) When classes of base classifiers (\eg, object classes and attribute classes) can be composed to form new composite classes (attribute-object classes), the composed classifier should support these composite classes as well. }
\label{fig:overview}
\end{figure}

\noindent \textbf{Union of Tasks.} When two classifiers $C_1$ and $C_2$ are trained independently  on two separate tasks (natural object classification and artificial object recognition respectively) as shown in Figure~\ref{fig:overview}(a), the resulting classifier should be able to perform classification across all classes that were supported in the base classifier (both natural and artificial objects). In Figure~\ref{fig:algebra}(b), we illustrate how prompt algebra functions in this specific use case. We interpolate prompts learned for natural object classification and artificial object classification to arrive at a prompt that performs  classification across both natural and artificial objects. \\

\noindent \textbf{Product of tasks.} If new object classes can be defined by composing classes of base classifiers together, the resulting classifier should perform well on these \textit{composite classes}. This situation arises when there exits two interpretations (in terms of annotation) for a given image. For the remainder of this paper, we refer such data as \textit{multi-view data}; where each \textit{view} refers to a unique interpretation.  For an example,  in Figure~\ref{fig:overview}(b), base classifiers are an object classifier and an attribute classifier. It is possible to define new classes by considering the product space of object and attribute classes. For example, \textit{young cat} is a composition of the concepts of \textit{young attribute} and \textit{cat object}. The resulting classifier should be able to recognize such composite concepts in addition to concepts supported by base classifiers. Prompt algebra can be used to generate a prompt (by interpolating base prompts) that recognizes objects, attributes and object-attribute pairs as shown in In Figure~\ref{fig:algebra}(c).  \\

In this paper,  we propose to select a prompt from the  convex hull of base prompts using a \textit{prompt algebraic} operation to support task compositionality. We argue that prompts should have a common grounding for them to gracefully compose and prevent any destructive interference. We encourage this property by constraining prompts to lie in a common subspace and introducing  regularization to ensure that pre-existing relationships in the pretrained vocabulary are preserved. We make following contributions in this paper:
 
 \begin{enumerate}
     \item We show that independently learned classifier models can be integrated together using \textit{prompt algebra} to support both union of task and  product of task compositionality.
     \item We propose a constrained prompt generation method based on loss regularization to improve performance of the composed classifier.

 \end{enumerate}

\section{Related Work}

    \noindent \textbf{Vision Language Models.} Recent works in both vision and language communities have explored the possibility of learning a joint vision-language embedding to support multi-modal tasks. Following the success of large language models \cite{bert,roberta,gpt}, earlier attempts have tried to reuse the structure of language models (such as BERT) to learn a joint embedding. VL-BERT\cite{vlbert} used faster R-CNN based object proposals to generate image tokens and combined them together with word tokens to learn a joint representation. On the other hand, VILBERT \cite{vilbert} used two separate language and vision towers to first obtain image and text features and then used a co-attention network to fuse features together. Most of later works adopted this strategy for learning.  CLIP \cite{clip} model optimized parameters of vision and language towers using a contrastive loss in the feature space. ALBEF model \cite{albef} added a secondary fusion network on top of vision-language features and used momentum distillation to learn representations. More recent works have used noisy-annotated data\cite{clip} and web-crawled data \cite{flamingo} to create powerful models that generalize well. These models have demonstrated strong zero-shot and few-shot recognition ability in multiple language-vision tasks\cite{clip, flamingo, align}.\\
    \noindent \textbf{Prompt Tuning.} Prompt tuning was first introduced in the NLP community to perform zero-shot inference with pretrained models . A \textit{text prompt}, in the traditional sense, is a set of words that describes the task desired from a network. This is now widely used with in-context-learning in NLP \cite{gpt3, palm}. It was later discovered that different prompts that are semantically similar can yield varying performance on ML tasks. Prompt engineering attempted to find the best prompt that produces optimum performance in a downstream task. However, it is both time and resource consuming as the optimization cannot be automated. Soft-prompt tuning \cite{liu-etal-2022-p,  xianglisaliprefixtuning,sprompt} attempts to replace prompt tuning with an optimization framework. Soft prompting optimizes over vectors in the word embedding space to find a set of representations that improves performance of the downstream task. It was later shown that prompting can be done even at a deeper level in a network to obtain better downstream task performance \cite{liupretrainprompt}. Several recent works have explored using soft prompt tuning in several NLP and vision tasks \cite{visualprompttuning}.\\
    \noindent \textbf{Product of Task Compositionality.} Several recent works have studied product of task compositionally in deep learning models. Given a multi-view dataset where each view is described by a finite set of classes, two views can be composed together to form a composite class. Object-attribute datasets have been commonly used to study the task compositionally of models. In this setting, objects and attributes serve as alternating views of the same image and the product space defined by \{attributes, objects\} becomes the composite-class space. Previous works have used transformation functions \cite{Misra_2017_CVPR, Lee_2022_CVPR}, late fusion networks\cite{https://doi.org/10.48550/arxiv.1905.05908} and graph neural networks \cite{ruis2021independent, naeem2021learning} to obtain better task compositionally. Recently, it was shown that prompt tuning with class specific prompts can obtain promising results in this task \cite{csp}. However, in all these methods, both views of the dataset is shown simultaneously to the learned classifier during training. In contrast, in this paper, we study the special case when two classifiers are trained on each view independently with partial annotation.\\ 
   \noindent \textbf{Continual Learning.} Continual learning is a learning paradigm where a single model is required a learn a series of tasks sequentially \cite{thengane2022continualclip, rebuffi-cvpr2017, dytox}. There exists three flavors of continual learning \cite{vandeVen2019ThreeSF}, namely task incremental learning,  domain incremental learning and class incremental learning (CIL). The setting we explore in this paper is similar in spirit to CIL with a notable difference: CIL typically involves retraining an existing model with data belonging to the new tasks (and possibly with replays from previous tasks) with the objective of performing well across union of tasks. In contrast, we explore ways to expand task support by combining  two independently trained classifiers on sub-tasks. 

\section{Background}

\subsection{Multi-class Classification with VLM}

Recent advances in VLM have resulted in models with better zero-shot classification capabilities \cite{clip, albef, align, flamingo}. In this paper we limit our discussion to VLM models with separate vision and language towers. Let us consider such a network as illustrated in Figure~\ref{fig:classification}. This network contains a vision encoder $e_v$ and a text encoder $e_t$. Such a pretrained network can be used to perform $c$ way classification.

Given a image $I$, first the vision tower is used to extract the vision feature $e_v(I)$. Then, class specific text feature $e_t(t_i)$ is calculated for a text of the form $t_i = [ t_{common} , c_i ]$, where $t_{common}$ is a class-agnostic prompt and $c_i$ is a class specific prompt for $i^\text{th}$ class. In our experiments we use class agnostic prompt $t_{common}$ to be the text "Image of a ". Class labels are used as class specific prompts $c_i$. Finally, the distance $d$ between the image feature  $e_v(I)$ and text features $e_t(t_i)$ are calculated. Cosine-distance  is commonly used as the distance metric. During training, the pretrained model is finetuned by considering image-text distance $d$ to be the class logit score and minimizing cross-entropy loss.

During inference, the class that produces lowest image-text feature distance $\arg\min_i d(e_v(I), e_t(t_i))$ is considered to be the model prediction. Zero-shot classification is trivial within this framework. A classifier can be converted into a zero-shot classifier by simply extending the list of prompts passed to the text encoder, by including zero-shot class labels.   

\subsection{Prompt Tuning}
Prompt Tuning is a recent advancement in Machine Learning \cite{lester-etal-2021-power} that allows output of a transformer based model to be influenced by a learnable set of token embeddings. A text-prompt of the form $t_i = [ t_{common} , c_i ]$ is extended to the form $t_i' = [ t_{common}, v, c_i ]$ by introducing a nominal token $v$. During training, embedding vector of the nominal token $e_t(v)$ is optimized with the aim of minimizing the model objective function. Prompt tuning is more parameter efficient, yet produces comparable performance to finetuning in various language and vision tasks \cite{liu-etal-2022-p}. Once trained, the prompt $v$ encodes information about the downstream task. For the remainder of the paper we refer to $v$ as the \textit{task prompt}.

\begin{figure}[h] 
\centering
\includegraphics[width=0.9\linewidth]{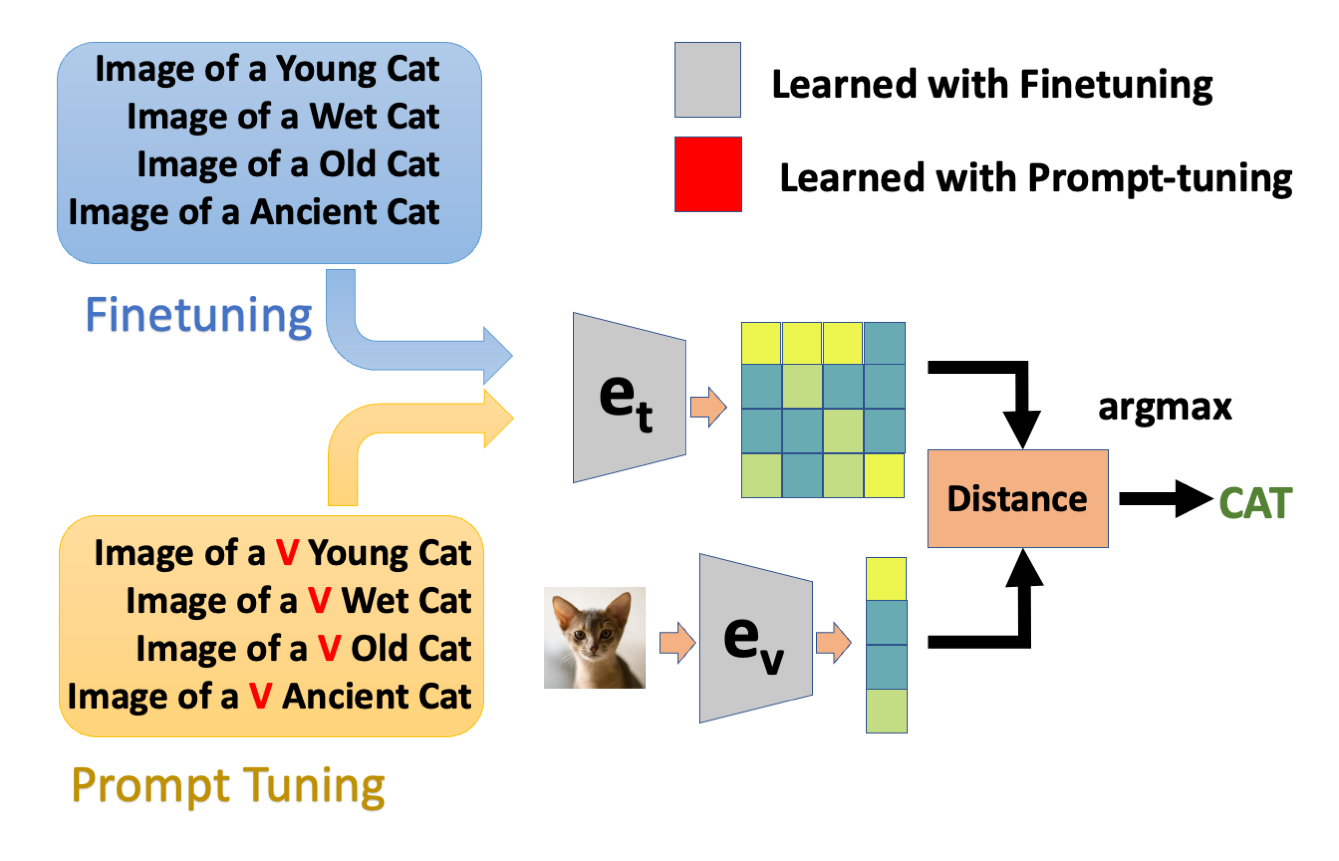} 
\caption{Overview of VLM based classification. Image is passed through the vision encoder $e_v$ to obtain image features. In finetuning, text prompts (in blue) are passed to the language encoder $e_t$ to get text features. Distance between Image features and text features is used to optimize parameters of $e_t$ and $e_v$ using a contrastive loss. In prompt tunning, text input (in yellow) is appended with a special token $v$. During training, all parameters except token embeddings of $v$ is kept fixed. Token embeddings of $v$ are optimized with respect to the contrastive loss same as before. }
\label{fig:classification}
\end{figure}

 A prompt-tuned VLM can be used for multi-class classification as shown in Figure~\ref{fig:classification}. The only difference from before is the incorporation of prompts when producing input to the text encoder (see the yellow colored input in Figure~\ref{fig:classification}). In this setup,   text features are calculated for a set of prompts $t_i = [ t_{common}, v, c_i ]$, where $v$ is the learned soft prompt.

\section{Composing Independent Classifiers} 
In this section, we describe our proposed solution for composing independently learned classifiers. First, we describe our choice of model backbone followed by details of prompt algebra. We end this section by introducing constrained prompt learning that encourages representations to have desired properties for prompt algebra.

\subsection{Model Architecture}
 We use a VLM equipped with prompt-tuning as our choice of model due to several reasons. VLM training and inference is free of class-specific layers. This enables easy integration of models trained on different tasks. A prompt is added as a part of the input to the network. Compared to other approaches that generate task specific architectural changes \cite{sung2022vladapter}, prompt tuning allows a direct way to manipulate model parameters by transforming prompts.
 For each task $j$, we train a class-agnostic prompt $ v_j$ to maximize model performance with respect to a given dataset. Specifically, for classification tasks, a text input to the text encoder is formed as $t_i = [ t_{common} , v_j  , c_i ]$. During training, parameter $ v_j $ is optimized by minimizing cross-entropy loss.
 
 \subsection{Prompt Algebra}
 Let $\mathbf{v} = [ v_0, v_1, \dots] $ be a set of prompts trained on a collection of datasets. The goal of prompt algebra is to create prompts for new tasks by  linearly combining the prompts $v_i$. That is, prompt algebra defines a linear transformation $f(\mathbf{v}) = \sum \theta_i v_i \forall \theta_i \geq 0$.

Pretrained language models and Vision-Language Models (VLM) exhibit properties of compositionality in the vocabulary space. In prompt tuning, a special token $v_j$ is learned for task $j$. Once trained, the special token $v_j$ modulates class label $c_i$ such that the prompt $t_i = [ t_{common} , v_j  , c_i ]$ produces highest correlation with image from class $i$ compared to images from other classes.  In this paper, we claim that soft prompts trained this way exhibit the same compositionality property their word-token counter part process.  i.e. if $e_t[v_A]$ and $e_t[v_B]$ are prompt embeddings learned for two tasks taskA and taskB, the prompt algebraic operation $\theta_A \cdot e_t[v_A]+ \theta_B \cdot e_t[v_B]$  results in a modulation that align images with labels of both tasks A and B (here, $\theta_A, \theta_B$ are constant weights). We use this property to compose independently trained prompt-tuning based classifiers. Prompts of the composite model are obtained by $\sum_{j} \theta_j  v_j $, where, $\theta_j$ is a weight associated with each task.

\subsection{Constrained Prompt Tuning.} In \Cref{sec:results}, we empirically show that prompt algebra is effective for prompts trained on certain datasets without any additional conditioning. However, this is not guaranteed to occur in every scenario due to multiple reasons. When task prompts are trained independently from each other, it's possible for prompts to appear in very different sub-spaces. There is also a possibility that prompts may interfere with each other as well as with other vocabulary tokens when this is the case. To avoid the former problem, we propose constraining prompt space to the lower dimensional subspace of the pre-trained vocabulary.

For the latter problem, we note that when VLMs soft prompts are trained on a downstream task, the prompt learns to attend to the task's labels of known classes. However, learned prompts have no incentive in learning a non-degenerate attention mechanism that will generalize well to unseen labels or contexts. Although this behaviour is acceptable when fine-tuning on a specific task, it may produce undesirable effects when prompts are combined and reused on new tasks. As a counter measure, we wish to regularize the learning to better ground the prompts in the pre-trained vocabulary tokens. This encourages the otimization to move along more meaningful subspaces and to learn more robust attention mechanisms. In this paper, we explore two forms of regularizations: multi-view regularization and class-agnostic regularization.\\

\noindent \textbf{Constraining prompt space.} In order to encourage compositionality, we aim to force prompts to reside in a common sub-space. We choose the lower dimensional subspace of the pre-trained vocabulary as our choice of the sub-space as it's a static reference. On the other hand, tokens in this space themselves are composible. Therefore, by making this choice, we hope soft prompts will gracefully compose with other vocabulary tokens as well.

Specifically, given the pretrained vocabulary $V$, we find the eigendecomposition of the co-relation matrix $V^TV$, and form a projection matrix $E$ using the eigenvectors of $V^TV$ that corresponds to largest $m$ eigenvalues. We compute $m$ in order to preserve a desired amount of spectral energy (90\% in our experiments). 
During both training and inference, each prompt token $v_j$ is projected to the lower dimensional subspace of $V$ using the projection operation $Pv_j$. The composite prompt vector now becomes $f(\mathbf{v}) = P \sum_i \theta_i \cdot v_i$.
In this process, the eigen decomposition happens only at the beginning of training. Therefore the additional processing overhead is fixed.\\


\newcommand{\viewA}{${\text{viewA}}$\xspace}
\newcommand{\viewB}{${\text{viewB}}$\xspace}
\newcommand{\Y}{\mathcal{Y}}

\begin{figure}[h] 
\centering
\includegraphics[width=0.4\textwidth]{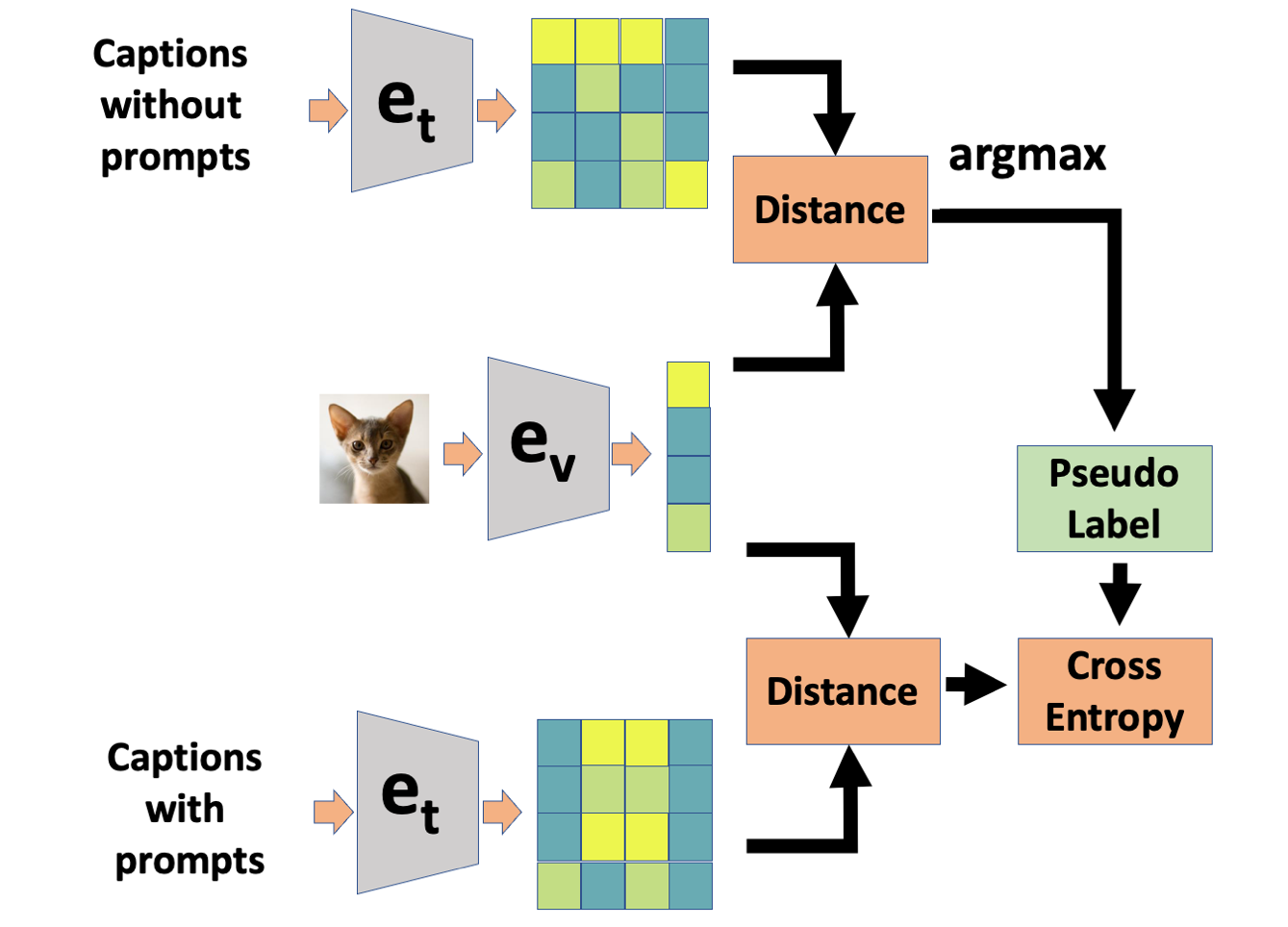} 
\caption{Overview of regularization. First, text prompts corresponding to some \textit{support classes} are passed  without prompts through the text encoder $e_t$. Distance between text feature and image features are used to determine the pseudo label among \textit{support classes}. Then,  text prompts of the same set of \textit{support classes} are passed through the text encoder after appending prompts. Cross entropy loss is used to ensure that text feature of the pseudo labelled \textit{support class}  has the lowest distance with respect to the same image feature. This regularization ensures pre-existing relations that existed in the pretrained vocabulary (with respect to {support classes}) are maintained during prompt tuning. }
\label{fig:reg}
\end{figure}

\noindent \textbf{Multi-view Regularization.} Multi-view regularization is a specialized regularization that can be applied when data has multiple views, i.e., when samples in the dataset have multiple annotations. For example, an image could be annotated both with an object category (e.g., cat) and an attribute (e.g., old). Let $\Y^A$ and $\Y^B$ be the label space corresponding to to possible views in the dataset. For an example, $\Y^A$ may be object labels and $\Y^B$ may be attribute annotations.  Suppose we are training a classifier on $\Y^A$. We use the following regularization procedure (outlined in Figure~\ref{fig:reg}):
\begin{enumerate}
\itemsep-0.2em 
    \item Let $(I, y^A_i)$ be an image-label pair from view A, for example $y^A_i = \texttt{cat}$.
    Let $\Y^B$ be the label space of the second view,  for example $\Y^B = \{$\texttt{young}, \texttt{black}, \texttt{old}, \texttt{ancient}$\}$.
    \item Sampling $k$ possible labels $y_l^B \in \Y^B$, we create text prompts of the form $t_l = [ t_\text{common}, y^B_l, y^A_i ]$, without the soft-prompt token. For an example, ``Image of a young cat'' could be one of the considered text prompts.
    \item  We search for the index $l^* = \arg \min_l d(e_v(I), e_t(t_l))$  that minimizes the distance between image and text features. $y^B_{l^*}$ is the pseudo-ground truth label.
    \item Then, a second set of text prompts are generated, this time including the soft-prompt token $t_l^v = [ t_\text{common},  v_j , y^B_l, y^A_i ]$. For the above example, text prompt now becomes ``Image of a $v$ Young Cat''.
    \item Finally, distance between image feature $e_v(I)$ and text-feature $e_t(t^v_l)$ is calculated for each class. Considering these distance values to be class logits, cross entropy loss is calculated using the pseudo-label $y^B_{l^*}$ as the target.
 \end{enumerate}   
The same regularization process can be applied for a classifier trained on $\Y^B$ , by using pre-trained model to find the pseudo object label  for $\Y^A$ in each given image.

\vspace{0.5em}

\noindent \textbf{Class-agnostic Regularization.} Class-agnostic regularization is a more generic form of regularization that can be applied to any type of a classifier. This form of regularization uses a pre-defined static set of generic classes as support classes to maintain grounding with the pre-trained model. For example, in our experiments we considered [material, animal, food, dress, place, vehicle, plant, object] to be the support class list. We construct text features for support classes without prompts, and obtain a pseudo ground truth annotation by considering highest correlation to a given image feature as shown in Fig~\ref{fig:reg}. Then, we generate text embeddings for the support set with soft prompts, and force the pseudo ground truth class to have the highest correlation with image features.

\begin{enumerate}
\itemsep-0.2em 
    \item Let $\Y^A = [ y^A_1, y^A_2, \dots , y^A_k]$ be the set of generic classes (support classes) identified. 
    \item For a given image $I$, consider $k$ sets of text prompts of the form $t_l = [ t_{common}, y^A_l ]$ not including the soft-prompt token.
    \item 
    We search for the index $l^* = \arg \min_l d(e_v(I), e_t(t_l))$ of the pseudo-ground truth label $y^A_{l^*}$ that minimizes the distance between image and text features.
    \item Then, a second set of text prompts are generated with the soft-prompt token of the form on $t_l^v = [ t_{common},  v_j , y_l^A]$.
    \item Finally, distance between image feature $e_v(I)$ and text-feature $e_t(t^v_l)$ is calculated for each class. Considering these distance values to be class logits, cross entropy loss is calculated with respect to the pseudo-attribute label $y^A_{l^*}$.
\end{enumerate}

In both forms of regularization, we record the response of the pre-trained model with respect to support classes and ensure, the same response is obtained when soft prompts are introduced. In order for these regularizations to be effective support class set should be different from the classifier classes.

\section{Experiments} \label{sec:results}

We evaluate the effectiveness of the proposed method on standard object classification datasets and object-attribute datasets. In all datasets, we identify two sets of super-classes and train two models for each super-class independently. In our evaluations we investigate how well a classifier composed with prompt algebra performs on these separate classification tasks. For Object Classification datasets we compare performance of vanilla prompt algebra with models trained with class-agnostic regularization. For object-attribute datasets, we consider both types of regularization.\\

\subsection{Model Details.} We use CLIP ViT-L/14 model as our backbone architechture. In all our experiments we used a single prompt for prompt tuning. We selected the eigenspace by considering 0.9\% of the spectral energy. In all experiments we gave equal weights to each prompt when performing prompt algebra. Each model was trained for 20 epochs with a dropout of 0.3 on prompt elements.  We used an effective batch size of 512. The model was optimized with SGD with a learning rate of 0.01. For all experiments we considered three trials of experiments and report average performance (with standard deviation within brackets).

\subsection{Performance on Attribute-Object Classification datasets}

We use multi-view datasets to analyze performance of the proposed method as a union of task classifier and product of task classifier. Both MIT-states\cite{mitstates} and UTZappos\cite{zappos} has object-attribute annotations for each data sample. We consider objects and attributes as two separate views of data for our analysis. The product space between these two views results in composite classes. For example, product of attribute \textit{young} and object \textit{cat} results in the composite class \textit{young cat}.

MIT-states contains images of naturally occurring objects where each object is described by an adjective. It contains 115 attribute classes and 245 object classes. In total, there are images belonging to 1962 composite classes. UT-Zappos contains images of shoes paired with attributes that describe the shoe. The dataset consists of 12 object classes and 16 attribute types and 118 composite object-attribute classes.  For both these datasets we used the splits used in \cite{csp} for bench-marking. In all datasets, only a selected few object-attribute combinations appear during training. These composite classes are named as seen-classes. Composite classes that appear only during testing are termed as unseen-classes. For object-attribute classification there is no notion of seen/unseen classes as all objects and attributes are observed during training.

For evaluation, we train classifiers for each modality for all datasets. During inference, we evaluate model performance on three criteria: test accuracy on the same modality, zero shot accuracy on cross modality, and zero shot performance on composite object-attribute task. We report best seen, best unseen and AUC metrics  \cite{csp} following closed-set protocol. For each method, we report performance obtained from base classifiers as well as composite classifier obtained with prompt algebra.

In Table~\ref{tbl:zappos}, we tabulate results obtained with UTZappos dataset. Interestingly, these results suggest that a object classifier trained on this dataset elevates attribute recognition performance over the CLIP baseline and vice versa. It should be noted attribute annotations were not used in any capacity when the object classifier is trained.  This phenomena can be observed even on MIT-states datasets as evident in Table~\ref{tbl:mitstates}. This could be due to the fact that object and attribute recognition is inter-related. Therefore, when a classifier is trained to learn features that are important for attribute classification, it  inadvertently learns features useful for object classification - there by improving zero-shot object recognition performance. 

Without any regularization, prompt algebra has produced a classifier with average performance on both views for UTZappos. Regularization further improves performance of this classifier. In particular, performance of the composite classifier on attribute recognition is better than the base attribute classifier. A similar gain can be observed in composite-class classification tasks (for seen and unseen pair accuracy). This is one of the rare cases where combining  classifiers of two views have resulted in a synergy. 

On the other hand, performance gain of the composite classifier is not significant for MIT-states dataset. Regularization has improved attribute and object classification performance by 3.25\% and 1.39\%  respectively. Seen and unseen object-attribute pair recognition has improved by approximately 2\%. This is partly due to the fact there exist a stronger correlation between objects and attributes in this dataset. For an example, an object classifier has a zero shot attribute recognition accuracy of 22.22\%; this is just 4.2\% worse than a specialized attribute classifier. \\

\begin{table*}[]
\centering
	\resizebox{0.8\linewidth}{!}{
\begin{tabular}{@{}ccccccc@{}}
\toprule
                  &                 & Attribute Accuracy    & Object Accuracy       & Seen Pair Accuracy    & Unseen Pair Accuracy  & AUC (Pair)           \\ \midrule
CLIP Zero Shot    &                 & 12.01                 & 48.76                 & 15.74                 & 49.23                 & 5.00                 \\ \hline
Prompt Tuning     & Obj. Classifier & 27.22 (2.07)          & 53.20 (0.82)          & 16.78 (0.83)          & 51.54 (1.40)          & 6.20 (0.39)          \\
                  & Att. Classifier & 43.93 (0.51)          & 34.18 (2.77)          & 10.62 (0.06)          & 51.40 (0.92)          & 4.18 (0.02)          \\
                  & Prompt Algebra  & 38.76 (3.04)          & 48.10 (0.81)          & 20.66 (0.76)          & 50.52 (1.16)          & 7.06 (0.67)          \\ \hline
MV-Regularization & Obj. Classifier & 13.77 (0.20)          & 48.26 (0.52)          & 22.29 (1.92)          & 50.77 (0.41)          & 7.73 (0.90)          \\
                  & Att. Classifier & 38.49 (0.61)          & 39.40 (1.96)          & 10.52 (0.45)          & 50.68 (0.34)          & 3.27 (0.10)          \\
                  & Prompt Algebra  & \textbf{47.48 (0.81)} & 47.70 (0.69)          & 22.68 (1.96)          & 50.91 (0.54)          & 8.13 (0.98)          \\ \hline
CA-Regularization & Obj. Classifier & 28.25 (1.61)          & 51.90 (1.35)          & 12.15 (0.76)          & 48.81 (0.32)          & 4.21 (0.18)          \\
                  & Att. Classifier & 40.60 (0.05)          & 30.30 (3.40)          & 10.90 (0.48)          & 52.75 (1.08)          & 4.44 (0.17)          \\
                  & Prompt Algebra  & 46.58 (0.49)          & \textbf{48.68 (0.58)} & \textbf{25.61 (2.14)} & \textbf{53.46 (1.64)} & \textbf{9.48 (1.45)} \\ \hline
\end{tabular}
}
\caption{Performance on UTZappos Dataset} \label{tbl:zappos} 
\end{table*}

\begin{table*}[]
\centering
	\resizebox{0.8\linewidth}{!}{
\begin{tabular}{@{}ccccccc@{}}
\toprule
                  &                 & Attribute Accuracy    & Object Accuracy       & Seen Pair Accuracy    & Unseen Pair Accuracy  & AUC (Pair)            \\ \midrule
CLIP Zero Shot    &                 & 15.78                 & 40.85                 & 30.50                 & 46.08                 & 11.00                 \\ \hline
Prompt Tuning     & Obj. Classifier & 22.22 (0.49)          & 52.40 (0.07)          & 32.54 (0.27)          & 48.16 (0.12)          & 12.59 (0.04)          \\
                  & Att. Classifier & 26.43 (0.49)          & 46.74 (0.57)          & 31.09 (0.30)          & 47.42 (0.73)          & 12.11 (0.16)          \\
                  & Prompt Algebra  & 21.57 (0.95)          & 48.72 (0.76)          & 31.27 (0.68)          & 46.83 (0.30)          & 11.68 (0.39)          \\ \hline
MV-Regularization & Obj. Classifier & 23.02 (0.03)          & 51.81 (0.02)          & 31.91 (0.03)          & 46.34 (0.06)          & 12.04 (0.01)          \\
                  & Att. Classifier & 26.00 (1.15)          & 48.45 (0.02)          & 33.26 (1.04)          & 48.36 (0.57)          & 13.09 (0.37)          \\
                  & Prompt Algebra  & 23.50 (0.50)          & 49.43 (0.81)          & 33.00 (0.56)          & \textbf{48.21 (0.24)} & 12.99 (0.08)          \\ \hline
CA-Regularization & Obj. Classifier & 22.85 (0.23)          & 52.29 (0.08)          & 32.56 (0.12)          & 48.36 (0.09)          & 12.77 (0.03)          \\
                  & Att. Classifier & 26.67 (0.67)          & 48.18 (0.16)          & 32.18 (0.59)          & 47.71 (0.09)          & 12.70 (0.14)          \\
                  & Prompt Algebra  & \textbf{24.32 (0.83)} & \textbf{50.11 (0.59)} & \textbf{33.51 (0.80)} & 48.05 (0.21)          & \textbf{13.01 (0.43)} \\ \hline
\end{tabular}
}

\caption{Performance on MIT-states Dataset} \label{tbl:mitstates}
\end{table*}

\subsection{Performance on Object Classification datasets.} We use CIFAR10\cite{cifar10} and CIFAR100\cite{cifar100} as benchmarks for object classification. In these datasets, we study how effective the proposed method is in creating a classifier that supports union of tasks. Specifically, we group all classes in these datasets into two super-classes: natural objects and artificial objects. CIFAR10 and CIFAR100 has 6(out of 10) and 31(out of 100) classes that qualify as natural objects. We train two classifiers independently: one on natural objects and the other on artificial objects. Results obtained for natural object classification, artificial object classification and overall classification performance is tabulated in  Tables~\ref{tbl:cifar10} and \ref{tbl:cifar100}. 

On CIFAR10 dataset, even the pretrained CLIP model exhibits very good zero shot performance. However  object classifiers (for both natural and artificial image classes) improves on zero-shot performances of the in-domain data-split by about 2\%. When a composite classifier is created with prompt algebra, it ends up producing marginally better overall classification performance (by 0.3\%). Adding Regularization had further improved overall classification performance by a 0.3\%. The table suggests that prompt algebra had produced a classifier that operates in-between a the two base classifiers. It's performance is better than what base classifiers have achieved for cross-split testing. 

CIFAR100 exhibits a similar trend to CIFAR10. Prompt algebra has improved best over-all classification performance by 2.6\% in this dataset. Regularization has decreased performance by about 0.5\% compared to the unregulated version. This could be due to the fact nominal classes used for regularization has higher correlation to CIFAR100 classes. In both datasets, performance of the composite classifier is within $1\%$ of the best base classifier. \\

\begin{table*}[]
\centering
	\resizebox{0.6\linewidth}{!}{

\begin{tabular}{@{}llccc@{}}
\\  \hline 
                  &                 & \multicolumn{1}{l}{Artificial Objects Accuracy} & \multicolumn{1}{l}{Natural Object Accuracy} & \multicolumn{1}{l}{Overall Accuracy} \\ \hline
CLIP Zero Shot    &                 & 96.55                                           & 94.67                                     & 94.99                                \\  \hline 
Prompt Tuning     & Artificial Obj. Classifier & 98.63 (0.07)                                    & 94.61 (0.47)                                & 96.04 (0.28)                         \\
                  & Natural Obj. Classifier  & 93.37 (0.79)                                    & 96.57 (0.12)                                & 95.09 (0.27)                         \\
                  & Prompt Algebra  & 97.51 (1.18)                                    & 95.77 (0.35)                                & 96.29 (0.69)                         \\  \hline 
CA-Regularization & Artificial Obj. Classifier & 98.54 (0.08)                                    & 93.79 (0.31)                                & 95.52 (0.16)                         \\
                  & Natural Obj. Classifier  & 96.71 (0.85)                                    & 96.77 (0.18)                                & 96.54 (0.38)                         \\
                  & Prompt Algebra  & \textbf{98.25 (0.25)}                           & \textbf{95.87 (0.10)}                       & \textbf{96.64 (0.12)}               \\  \hline 
\end{tabular}}

\caption{Results on CIFAR10 Dataset} \label{tbl:cifar10}
\end{table*}

\begin{table*}[]
\centering
	\resizebox{0.6\linewidth}{!}{
\begin{tabular}{@{}llccc@{}}
\\ \hline
                  &                 & \multicolumn{1}{l}{Artificial Objects Accuracy} & \multicolumn{1}{l}{Natural Object Accuracy} & \multicolumn{1}{l}{Overall Accuracy} \\ \hline
CLIP Zero Shot    &                 & 65.43                                           & 74.71                                       & 65.28                                \\ \hline
Prompt Tuning     & Artificial Obj. Classifier & 78.27 (0.34)                                    & 81.47 (0.20)                                & 77.93 (0.46)                         \\
                  & Natural Obj. Classifier  & 73.67 (0.39)                                    & 88.02 (0.66)                                & 78.93 (0.47)                         \\
                  & Prompt Algebra  & \textbf{76.99 (0.28)}                           & \textbf{86.83 (0.86)}                       & \textbf{78.93 (0.47)}                \\ \hline
CA-Regularization & Artificial Obj. Classifier  & 78.64 (0.70)                                    & 83.73 (1.42)                                & 78.91 (0.78)                         \\
                  & Natural Obj. Classifier & 74.67 (0.16)                                    & 88.30 (0.21)                                & 77.53 (0.03)                         \\
                  & Prompt Algebra  & 76.89 (0.15)                                    & 86.52 (0.23)                                & 78.77 (0.10)                        \\ \hline
\end{tabular}}
\caption{Results on CIFAR100 Dataset}
\label{tbl:cifar100}
\end{table*}

\begin{table}[]
\centering
	\resizebox{0.6\linewidth}{!}{
\begin{tabular}{@{}lll@{}} \\ \hline
                         & \multicolumn{2}{c}{10 steps} \\
                         & Average          & Last      \\ \hline
DER (w/o P) \cite{der}             & 75.36 (0.36)     & 65.22     \\
DER \cite{der}                     & 74.64 (0.28)     & 64.35     \\
DyTox \cite{dytox}                    & 67.33 (0.02)     & 51.68     \\
DyTox+  \cite{dytox}                  & 74.1 (0.10)      & 62.34     \\ 
Continual CLIP \cite{thengane2022continualclip}          & 75.17 (-)        & 66.72     \\ \hline
Continual CLIP (our run) & 77.48 (0.09)     & 65.28     \\
Prompt Algebra           & 85.29 (0.05)   & 80.10   \\ 

Prompt Algebra(CA)          & 84.37 (0.07)   & 76.87   \\ \hline
\end{tabular}}
\caption{Continual learning results on CIFAR100 Dataset}
\label{tbl:cil}
\end{table}

\subsection{Continual Learning Performance.} Class Incremental Learning (CIL) is different from our problem setting. Both settings set to solve classification problem in a union of tasks setting. However, information is shared between two tasks in CIL, whereas there is no information shared between models in our setting. Nevertheless, we tried to apply prompt algebra on a CIL task to test its applicability.

For this experiment we use CIFAR100-10 step protocol. In this protocol a model is initially trained on 10 classes. Then, the model encounters 10 new classes at a time - where it is expected to adapt new data while maintaining it's expertise on older classes. Performance is evaluated by considering average performance across all splits and classification accuracy at the end of training (denoted as final).

We used prompt algebra to solve CIL task by training a new prompt for  every split of 10 classes. We combined trained prompts (with equal weight) to obtain a composite classifier.  In Table~\ref{tbl:cil}, we compare performance obtained with out method with recent state-of-the as reference.  Here we show that prompt algebra out-performs continual CLIP\cite{thengane2022continualclip} by more than 7\% in average. It should be noted that regularization resulted in slightly worse performance than vanilla prompt algebra for this task. \\

\subsection{Ablation Study}

In this section, we analyze design choices made in the proposed method. For a ablation of spectral energy, number of prompts and weight of the regularization please refer the supplementary materials. 

\noindent \textbf{Impact of Regularization.} In Table~\ref{tbl:ablation} we analyzed the impact of each component of regularization with a single trial of experiments on UTZappos dataset. Note that performance numbers of ablation study is different from results in the main section. According to Table~\ref{tbl:ablation}, adding MV-regularization improved average recognition performance by over 3\%. Training classifier with Eigen projection has resulted an improvement over 5\%.  When both forms of regularization are added average model performance increases by almost 6\%. This study shows that each regularization step positively impacts model performance and that they can be used together when training a model to a good effect.

\begin{figure}[h] 
\centering
\includegraphics[width=1.0\linewidth]{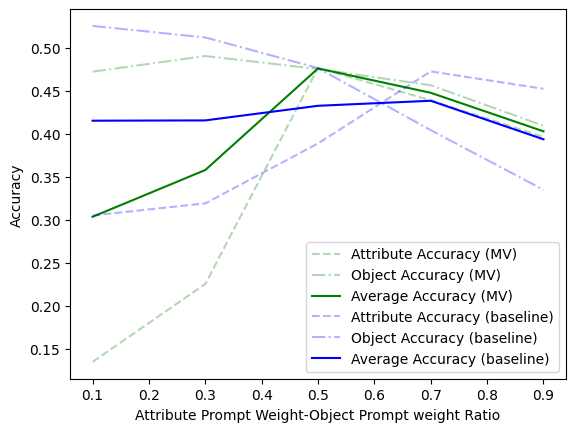} 
\caption{Impact of using different weights to combine object-attribute classifiers using prompt algebra on UTZappos dataset.}
\label{fig:weights}
\end{figure}

\noindent \textbf{Impact of Prompt Weight Coefficients.} In the main result section we consider equal weights when using prompt algebra on trained prompts. In this sub-section we explore the impact of using a different weight combination on UTZappos dataset. In Figure~\ref{fig:weights} we have plot object recognition performance of object classifier, attribute classifier and  composite classifier  for different combinations of weights. This figure suggests that performance curves are not always linear.  For an example, prompt algebra for regularized model has a peak around 0.5 in the performance curve. This indicates that it's possible to find synergies in composite models if weights are manually tuned.


\begin{table}[]
\centering
	\resizebox{1.0\linewidth}{!}{
\begin{tabular}{@{}cccccc@{}}
\begin{tabular}[c]{@{}c@{}}Eigen Space \\ Projection\end{tabular} & \begin{tabular}[c]{@{}c@{}}MV-\\ Regularization\end{tabular} & Model           & Attribute Accuracy & Object Accuracy & Average        \\ \hline 
\multirow{3}{*}{\xmark}                                                & \multirow{3}{*}{\xmark}                                           & Obj. Classifier & 14.76              & 40.01           & 27.39          \\
                                                                  &                                                              & Att. Classifier & 43.21              & 37.75           & 40.48          \\
                                                                  &                                                              & Prompt Algebra  & 42.79              & 40.56           & 41.68          \\ \hline 
\multirow{3}{*}{\cmark}                                                & \multirow{3}{*}{\xmark}                                           & Obj. Classifier & 28.17              & 52.20           & 40.19          \\
                                                                  &                                                              & Att. Classifier & 41.73              & 31.26           & 36.50          \\
                                                                  &                                                              & Prompt Algebra  & 45.92              & 48.76           & 47.34          \\ \hline 
\multirow{3}{*}{\xmark}                                                & \multirow{3}{*}{\cmark}                                           & Obj. Classifier & 13.80              & 45.23           & 29.51          \\
                                                                  &                                                              & Att. Classifier & 39.12              & 35.66           & 37.39          \\
                                                                  &                                                              & Prompt Algebra  & 45.92              & 44.03           & 44.97          \\ \hline 
\multirow{3}{*}{\cmark}                                                & \multirow{3}{*}{ \cmark }                                           & Obj. Classifier & 12.90              & 47.36           & 30.13          \\
                                                                  &                                                              & Att. Classifier & 40.73              & 35.18           & 37.95          \\
                                                                  &                                                              & Prompt Algebra  & 48.11              & 47.15           & \textbf{47.63} \\ \hline 
\end{tabular}
}
\caption{Ablation study on different components of regularization.} \label{tbl:ablation}
\end{table}

\section{Conclusion}
Our experimental results suggest that prompts trained with vanilla prompt tuning on a single view in a multi-view dataset (such as object-attribute datasets), elevates zero-shot performance of the second view. Once prompts are combined through algebraic operations, the composite classifier exhibits on par performance on both views as well as the composite view with respect to the base classifier. This observation continues to hold true even when classifiers with different class support are combined with prompt algebra (as in the case of CIFAR10/100 experiments). We observe that the proposed constrained prompt learning scheme further boosts performance of the composite classifier. In the future, we hope to extend this method to multiple prompts and explore an automated weight assignment mechanism for combining different prompts.

{\small
\bibliographystyle{ieee_fullname}

}

\end{document}